%% file: main.tex
\documentclass[10pt,twocolumn,letterpaper]{article}

\usepackage[pagenumbers]{cvpr}


\definecolor{cvprblue}{rgb}{0.21,0.49,0.74}
\usepackage[pagebackref,breaklinks,colorlinks,allcolors=cvprblue]{hyperref}

\def\confName{CVPR}
\def\confYear{2027}

\usepackage{hyperref}
\usepackage{url}
\usepackage{booktabs}
\usepackage[table]{xcolor}
\usepackage{graphicx}
\usepackage{threeparttable}
\usepackage{multirow}
\usepackage{bm}
\usepackage{cuted}
\usepackage{caption}

\newcommand{\vv}{\mathbf{v}}
\newcommand{\vx}{\mathbf{x}}

\newcommand{\vc}{\mathbf{c}}
\newcommand{\R}{\mathbb{R}}
\newcommand{\vepsilon}{\boldsymbol{\varepsilon}}

\definecolor{sectiongray}{gray}{0.90}
\definecolor{oursgreen}{RGB}{225,248,225}

\newcommand{\ours}{MS-pDiT}

\title{Scale as a Modality: Multi-Scale Pixel-Space Diffusion Transformers}

\author{Ruibo Chen$^{1,2,}\thanks{This work was done while Ruibo Chen was interning at TikTok.}$\ ,\ Kaishen Wang$^{2}$, Tianyi Xiong$^2$,\ Chenxi Liu$^2$,\ Yanshuo Chen$^2$, \\
Heng Huang$^{2,}\thanks{Corresponding Authors}$\ ,\ Jian Du$^{1,\dagger}$ \\
$^1$TikTok, $^2$University of Maryland, College Park
}

\begin{document}
\maketitle

\begin{strip}
    \centering
    \includegraphics[width=0.98\linewidth]{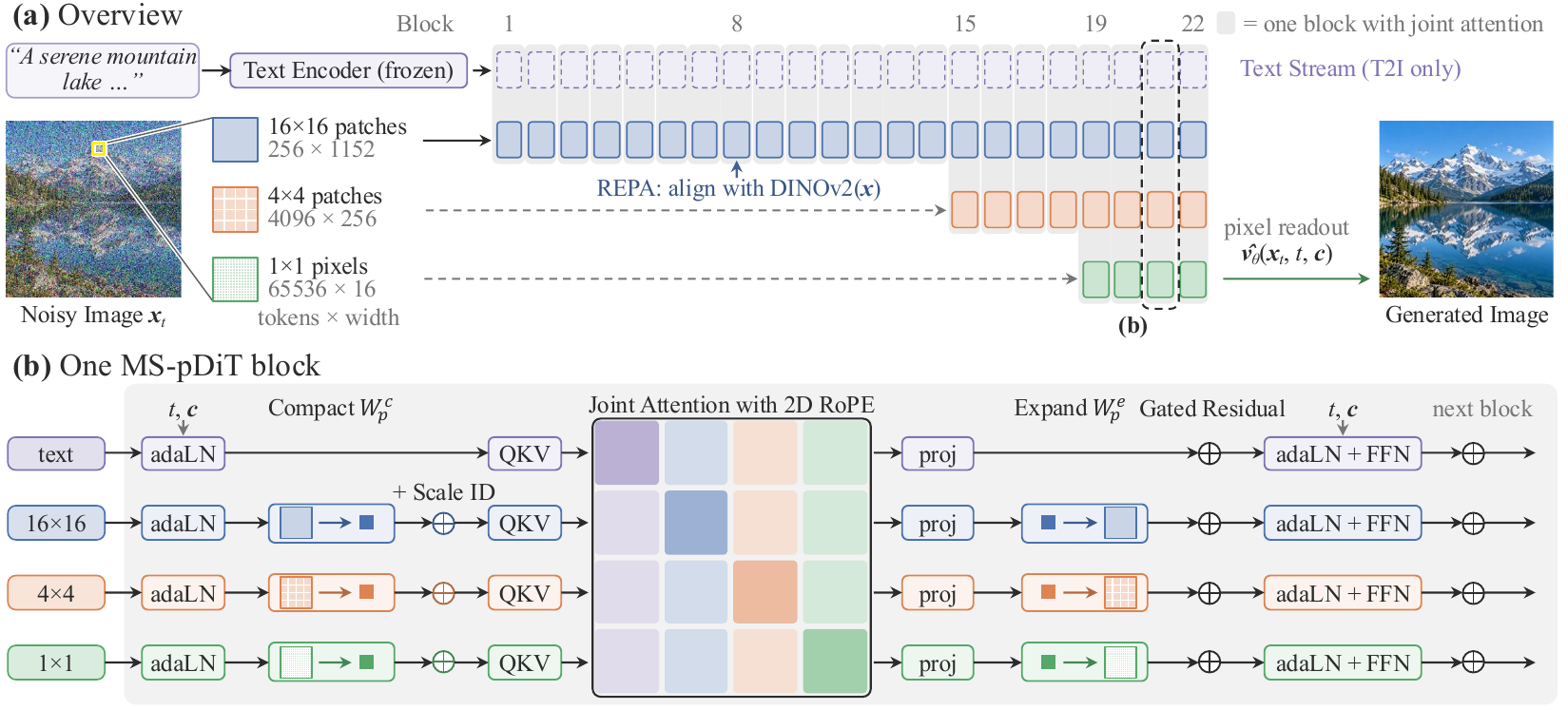}

    \captionsetup{hypcap=false}
    \captionof{figure}{
        Overview of MS-pDiT.
        (a) The same noisy image is tokenized at multiple
        spatial scales and processed by scale-specific
        transformer streams, with finer streams introduced
        only in later layers. An optional text stream enables
        text-to-image generation, while the final prediction
        is read out directly from the \(1\times1\) pixel stream.
        (b) In each MS-pDiT block, active image streams are
        compacted onto a shared spatial grid, augmented with
        scale identities, and coupled through MM-DiT-style
        joint attention with shared 2D RoPE, enabling
        bidirectional information exchange across scales.
        The attended features are then expanded back to
        their native resolutions and processed by
        stream-specific residual and feed-forward layers.
    }
    \captionsetup{hypcap=true}
    \label{fig:overview}
\end{strip}

\begin{abstract}
Latent diffusion models rely on pretrained autoencoders that introduce lossy compression and constrain spatial granularity. Pixel-space diffusion avoids this bottleneck and enables flexible tokenization at multiple spatial scales, yet existing approaches largely adopt two-level coarse-to-fine architectures with limited cross-scale interaction. We introduce \ours{}, a multi-scale pixel-space diffusion transformer that treats spatial scale as a modality. \ours{} tokenizes the same noisy image at multiple patch sizes, processes each scale in a dedicated transformer stream, and enables bidirectional information exchange through compacted MM-DiT-style joint attention, with attention cost independent of the finest patch size. This design establishes the number of scales as a new architectural scaling dimension. On ImageNet $256\times256$, increasing the number of scales from two to five consistently improves FID from 4.39 to 2.31, 2.26, and 2.19, while adding an intermediate scale substantially outperforms increasing depth under comparable parameter budgets. Further analysis highlights the importance of bidirectional cross-scale interaction and late introduction of fine-grained streams. \ours{} achieves FID scores of 1.43 and 1.44 on ImageNet $256\times256$ and $512\times512$, respectively, and naturally extends to text-to-image generation, achieving 0.86 on GenEval and 83.9 on DPG-Bench.

\end{abstract}

\section{Introduction}

Latent diffusion models have become the dominant paradigm for image generation~\citep{rombach2022high,peebles2023scalable,esser2024scaling}, with pretrained autoencoders compressing images into compact latent representations before generative modeling. This efficiency, however, introduces several inherent limitations. The autoencoder incurs lossy compression, separates representation learning from generative training, and fixes the spatial granularity at which the diffusion model operates~\citep{yao2025reconstruction,zheng2025diffusion,yu2026pixeldit}. Pixel-space diffusion eliminates this bottleneck by modeling RGB images directly~\citep{li2026back,yu2026pixeldit,ma2026deco,chen2025pixelflow}. Beyond avoiding reconstruction error and enabling end-to-end generative modeling, pixel space offers an important architectural advantage: \emph{flexible tokenization}. Unlike latent representations whose spatial granularity is determined by a pretrained autoencoder, raw pixels can be partitioned at arbitrary patch sizes. The same image can therefore be represented simultaneously at multiple granularities, allowing coarse tokens to capture global structure while finer tokens preserve local appearance and detail.

Existing pixel-space diffusion transformers~\citep{yu2026pixeldit,ma2026deco,dip2026,wang2026pixnerd,he2026hyperdit} exploit this flexibility only partially, typically employing only two granularities with a coarse pathway capturing global semantics, while a fine pathway reconstructs or refines local details. Moreover, information is primarily propagated in a coarse-to-fine direction, leaving the fine representation unable to influence the evolving coarse representation.

We address these questions with \ours{}, a multi-scale pixel-space diffusion transformer that treats \emph{scale as a modality}. Given the same noisy image, \ours{} independently tokenizes it at multiple patch sizes and assigns each granularity a dedicated transformer stream. To enable efficient interaction among streams with vastly different token counts, we compact their representations onto a shared spatial grid and couple them through a single MM-DiT-style joint attention operation~\citep{esser2024scaling}. This formulation enables unrestricted bidirectional information exchange across scales while keeping the attention cost independent of the finest patch size. Fine-grained streams are introduced only in later layers, whereas the coarse stream spans the full network to establish global structure before joint refinement. The final prediction is produced directly from the $1\times1$ stream, without an image autoencoder or decoder.

A central consequence of this design is that the \emph{number of scales becomes a new architectural scaling dimension}. Experiments indicate that scaling across spatial granularities provides a distinct and substantially more effective use of model capacity than conventional depth scaling. Our analysis further reveals two important principles for multi-scale pixel modeling. First, cross-scale interaction should be bidirectional: restricting attention to coarse-to-fine information flow greatly degrades performance, demonstrating that fine-grained representations provide useful feedback to the coarse stream. Second, fine-scale computation is most effective when introduced in later layers, compared to intermediate ones. With this design, \ours{} achieves 1.43 FID on ImageNet $256\times256$ and 1.44 FID at $512\times512$, establishing strong performance among pixel-space diffusion models. The same formulation naturally extends to text-to-image generation by treating text as an additional modality, achieving 0.86 on GenEval and 83.9 on DPG-Bench.

Our contributions are summarized as follows:
\begin{itemize}
    \item We identify flexible tokenization as a fundamental architectural advantage of pixel-space diffusion and propose \ours{}, a multi-scale pixel-space diffusion transformer that treats spatial scales as parallel modalities and enables bidirectional cross-scale interaction through compacted joint attention.

    \item We establish the number of scales as a new architectural scaling dimension. Controlled studies show that introducing intermediate granularities is substantially more effective than allocating comparable capacity to additional depth, and reveal bidirectional cross-scale interaction and late fine-stream computation as key design principles.

    \item Extensive experiments demonstrate strong performance across both class-conditional and text-to-image generation. \ours{} achieves FID scores of 1.43 and 1.44 on ImageNet at $256\times256$ and $512\times512$, respectively, and naturally extends to text conditioning, attaining 0.86 on GenEval and 83.9 on DPG-Bench.
\end{itemize}

\section{Related Work}
\label{sec:related}

\subsection{Latent and Pixel-Space Diffusion}

Diffusion transformers are predominantly trained on the latents of a
pretrained
autoencoder~\citep{rombach2022high,peebles2023scalable,ma2024sit,esser2024scaling}.
We instead remove the autoencoder entirely.  Pixel-space diffusion first
reached high resolutions through cascaded
models~\citep{ho2022cascaded,saharia2022photorealistic} or U-Nets with
aggressive downsampling~\citep{hoogeboom2023simple,hoogeboom2024simpler}, and
single-stage pixel transformers have recently become competitive through
$x$-prediction, self-supervised pretraining, and perceptual
losses~\citep{li2026back,ma2026pixelgen,chen2025pixelflow}.  Most
closely related to our work are PixelDiT~\citep{yu2026pixeldit}, DeCo~\citep{ma2026deco}, DiP~\citep{dip2026}, PixNerd~\citep{wang2026pixnerd}, and
HyperDiT~\citep{he2026hyperdit},
which divide the network into two granularities with strictly coarse-to-fine
conditioning, realized through a fine-level decoder, prediction head, or
cross-attention branch.  PixelU replaces the decoder with skip connections but
retains a two-resolution hierarchy~\citep{guo2026pixelu}.  In contrast,
\ours{} assigns a complete transformer stream to each of several
granularities and couples them bidirectionally. See Appendix~\ref{app:related} for an extended discussion.

\subsection{Multi-scale Generation}

Generative models typically exploit scale \emph{sequentially}: they first
generate a low-resolution image and subsequently refine it through cascades,
nested resolutions, or next-scale
prediction~\citep{ho2022cascaded,teng2024relay,gu2024matryoshka,tian2024visual}.
In \ours{}, all streams observe the same image at the same noise level within a
single forward pass; scale thus determines \emph{how the signal is tokenized}
rather than \emph{when it is generated}.

\section{Method}
\label{sec:method}

In this section, we introduce MS-pDiT, a multi-scale pixel-space diffusion transformer that models the same noisy image at multiple spatial granularities. As illustrated in Figure~\ref{fig:overview}, we first construct scale-specific pixel streams, then couple them through compacted joint attention to enable efficient bidirectional cross-scale interaction, and finally decode the prediction directly from the finest pixel stream. We further describe how the same formulation naturally incorporates text as an additional modality for text-to-image generation.

\subsection{Preliminaries}
\label{sec:preliminaries}

We model RGB pixels directly using linear interpolation (rectified-flow)
probability path~\citep{liu2023flow,lipman2023flow,albergo2023stochastic}.  Let
$\vx\in\R^{H\times W\times 3}$ denote a clean image scaled to $[-1,1]$,
$\vepsilon\sim\mathcal{N}(\mathbf{0},\mathbf{I})$ a noise sample of the same
shape, and $\vc$ the condition.  For
$t\in[0,1]$, the noisy image and its velocity are
\begin{equation}
    \vx_t = t\,\vx + (1-t)\,\vepsilon,
    \qquad
    \vv = \frac{\mathrm{d}\vx_t}{\mathrm{d}t} = \vx - \vepsilon,
    \label{eq:interpolant}
\end{equation}
so that $\vx_0=\vepsilon$ and $\vx_1=\vx$.  A network $\vv_\theta(\vx_t,t,\vc)$
is trained to regress this velocity with the flow-matching objective
\begin{equation}
    \mathcal{L}_{\mathrm{FM}}(\theta)
    =\mathbb{E}_{(\vx,\vc),\,\vepsilon,\,t}
      \Bigl[\bigl\lVert
      \vv_\theta(\vx_t,t,\vc)-(\vx-\vepsilon)
      \bigr\rVert_2^2\Bigr].
    \label{eq:flow_matching}
\end{equation}
The minimizer of Equation~\ref{eq:flow_matching} is the conditional expectation
$\vv^\star(\vx_t,t,\vc)=\mathbb{E}[\vx-\vepsilon\mid\vx_t,t,\vc]$, and integrating
the ODE $\mathrm{d}\vx_t/\mathrm{d}t=\vv^\star(\vx_t,t,\vc)$ from $t=0$ to $t=1$
transports $\mathcal{N}(\mathbf{0},\mathbf{I})$ to the data distribution
$p(\vx\mid\vc)$~\citep{lipman2023flow,liu2023flow}.  In contrast to
JiT~\citep{li2026back}, which predicts $\vx$ and converts the prediction into a
velocity, we regress $\vv$ directly with a time-independent loss weight, following~\citep{yu2026pixeldit}.

\begin{table*}[t]
    \centering
    \caption{
        Class-conditional generation on ImageNet $256\times256$.  All
        metrics are computed from 50K samples generated with classifier-free
        guidance.  ``Params'' denotes the inference parameter count
        (generator + VAE/RAE decoder), and NFE denotes the number of function
        evaluations.
        \textsuperscript{\ensuremath{\dagger}}~Results reported in the
        original paper.
        \textsuperscript{\ensuremath{\ddagger}}~Results reproduced with the
        official implementation, using post-modulation for improved training
        stability.
        Bold denotes the best result within each group.
    }
    \label{tab:merged_imagenet_256}

    \setlength{\tabcolsep}{10pt}
    \renewcommand{\arraystretch}{1.08}

    \begin{tabular}{lcccccccc}
        \toprule
        \textbf{Method}
        & \textbf{Params}
        & \textbf{Epochs}
        & \textbf{NFE}
        & \textbf{FID$\downarrow$}
        & \textbf{IS$\uparrow$}
        & \textbf{sFID$\downarrow$}
        & \textbf{Prec.$\uparrow$}
        & \textbf{Rec.$\uparrow$} \\
        \midrule

        \rowcolor{sectiongray}
        \multicolumn{9}{l}{\textit{Latent-space diffusion}} \\

        DiT-XL/2~\citep{peebles2023scalable}
        & 675{+}49M & 1400 & $250{\times}2$
        & 2.27 & 278.2 & 4.60 & \textbf{0.83} & 0.57 \\

        SiT-XL/2~\citep{ma2024sit}
        & 675{+}49M & 1400 & $250{\times}2$
        & 2.06 & 277.5 & \textbf{4.50} & 0.80 & 0.64 \\

        REPA-XL/2~\citep{yu2025representation}
        & 675{+}49M & 800 & $250{\times}2$
        & 1.42 & 305.7 & 4.70 & 0.80 & 0.64 \\

        DDT-XL/2~\citep{wang2025ddt}
        & 675{+}49M & 400 & --
        & 1.26 & \textbf{310.6} & -- & 0.78 & \textbf{0.67} \\

        RAE-XL/2~\citep{zheng2025diffusion}
        & 839{+}415M & 800 & --
        & \textbf{1.13} & 262.6 & -- & 0.78 & \textbf{0.67} \\

        \midrule
        \rowcolor{sectiongray}
        \multicolumn{9}{l}{\textit{Pixel-space diffusion}} \\

        ADM-G~\citep{dhariwal2021diffusion}
        & 554M & 400 & 250
        & 4.59 & 186.7 & 5.25 & \textbf{0.82} & 0.52 \\

        RIN~\citep{jabri2023scalable}
        & 410M & 480 & --
        & 3.42 & 182.0 & -- & -- & -- \\

        VDM++~\citep{kingma2023understanding}
        & 2B & -- & --
        & 2.12 & 278.1 & -- & -- & -- \\

        PixelFlow-XL/4~\citep{chen2025pixelflow}
        & 677M & 320 & $120{\times}2$
        & 1.98 & 282.1 & 5.83 & 0.81 & 0.60 \\

        PixNerd-XL/16~\citep{wang2026pixnerd}
        & 700M & 320 & $100{\times}2$
        & 1.95 & 300.0 & 4.54 & 0.80 & 0.60 \\

        JiT-G/16~\citep{li2026back}
        & 2B & 600 & $100{\times}2$
        & 1.82 & 292.6 & -- & -- & -- \\

        PixelGen-XL/16~\citep{ma2026pixelgen}
        & 676M & 160 & $100{\times}2$
        & 1.83 & 293.6 & -- & 0.79 & 0.63 \\

        DiP-XL/16~\citep{dip2026}
        & 631M & 600 & $100{\times}2$
        & 1.79 & 281.9 & 4.59 & 0.80 & 0.63 \\

        DeCo-XL/16~\citep{ma2026deco}
        & 682M & 600 & $100{\times}2$
        & 1.69 & 304.0 & 4.59 & 0.79 & 0.63 \\

        PixelDiT-XL\textsuperscript{\ensuremath{\dagger}}~\citep{yu2026pixeldit}
        & 797M & 320 & $100{\times}2$
        & 1.61 & 292.7 & 4.68 & 0.78 & 0.64 \\

        PixelDiT-XL\textsuperscript{\ensuremath{\ddagger}}
        & 797M & 320 & $100{\times}2$
        & 1.66 & 292.7 & 4.89 & 0.78 & 0.64 \\

        PixelU-H/16~\citep{guo2026pixelu} & 1.2B & 600 & $100{\times}2$
        & 1.63 & 305.9 & 5.04 & 0.79 & 0.64 \\

        \rowcolor{oursgreen}
        \textbf{\ours-L}
        & 788M & 600 & $100{\times}2$
        & 1.55 & 305.9 & 4.42 & 0.78 & \textbf{0.65} \\

        \rowcolor{oursgreen}
        \textbf{\ours-XL}
        & 963M & 600 & $100{\times}2$
        & 1.50 & \textbf{308.7} & 4.31
        & 0.79 & \textbf{0.65} \\

        \rowcolor{oursgreen}
        \textbf{\ours-H}
        & 1.4B & 360 & $100{\times}2$
        & \textbf{1.43} & 304.6 & \textbf{4.13}
        & 0.79 & \textbf{0.65} \\

        \bottomrule
    \end{tabular}
    
\end{table*}

\subsection{Multi-Scale Pixel Streams}
\label{sec:streams}

Let $\mathcal{P}$ denote the set of patch sizes. For each $p\in\mathcal{P}$, the same noisy image $\vx_t$ is partitioned into $p\times p$ patches and embedded using a scale-specific linear projection, yielding $N_p=HW/p^2$ tokens $\bm{X}_p\in\R^{N_p\times d_p}$. Each stream has its own normalization layers, attention projections, and feed-forward network. Importantly, these streams form parallel representations of the same image rather than successive stages of a coarse-to-fine hierarchy: each is embedded directly from $\vx_t$, with no stream derived from another. Consequently, fine streams preserve local structure, that is lost under coarse tokenization and cannot be recovered from the coarse stream alone. The predicted velocity is read out directly from the $1\times1$ stream, eliminating the need for patch-level upsampling at the output. Each stream has its own embedding dimension $d_p$, with finer streams kept narrower to accommodate their substantially larger token counts. Our default class-conditional model uses $\mathcal{P}=\{1,4,16\}$ with $(d_1,d_4,d_{16})=(16,256,1152)$. Complete width configurations for all models are provided in Appendix~\ref{app:hparams}.

Streams need not span the full network depth. Let $L_p$ denote the depth of stream $p$ and $L=\max_p L_p$ the total network depth. We right-align the streams so that stream $p$ is active only in the final $L_p$ blocks, with each block instantiating parameters only for its active streams. Most of the network depth is allocated to the coarse stream, which captures global semantics, consistent with prior pixel-space architectures that concentrate capacity in the patch-level pathway~\citep{yu2026pixeldit,ma2026deco}. Our default model uses $(L_1,L_4,L_{16})=(4,8,22)$: the $16\times16$ stream spans all 22 blocks, while the $4\times4$ and $1\times1$ streams are introduced only in the final eight and four blocks, respectively. Section~\ref{sec:anchor_analysis} compares this late anchoring strategy with an alternative that places the fine-stream blocks at intermediate depths~\citep{he2026hyperdit}.

\subsection{Compacted Multi-Scale Joint Attention}
\label{sec:compaction}

Direct attention over fine-grained streams is prohibitively expensive. We therefore perform attention on a shared coarser grid. Let the attention patch size $a$ be a common multiple of all $p\in\mathcal{P}$, with $a=16$ by default. Similar to~\citet{yu2026pixeldit}, for each stream $p$, the $(a/p)^2$ tokens within each $a\times a$ region are concatenated along the channel dimension and projected to a shared width $D$:
\begin{equation}
    \widetilde{\bm{X}}_p
    = \operatorname{concat}_{(a/p)^2}\!\left(\bm{X}_p\right)\bm{W}^{\mathrm{c}}_p
    \in\mathbb{R}^{M\times D},
    \label{eq:scale_compaction}
\end{equation}
where $M=HW/a^2$ and $\bm{W}^{\mathrm{c}}_p\in\mathbb{R}^{(a/p)^2 d_p\times D}$.
Thus, each stream contributes exactly $M$ compact tokens, with tokens at the same spatial position corresponding to the same $a\times a$ image region. After attention, a scale-specific expansion projection $\bm{W}^{\mathrm{e}}_p\in\R^{D\times(a/p)^2d_p}$ maps each compact token back to $(a/p)^2$ native-resolution tokens. As a result, attention operates on a sequence length determined by $M$ and $|\mathcal{P}|$, independent of the finest patch size, while each stream maintains its native-resolution representation for normalization and feed-forward computation.

All image streams share the same 2D rotary position embeddings~\citep{su2024roformer} over the $H/a\times W/a$ attention grid. This shared coordinate system establishes explicit cross-scale correspondence: compact tokens from different streams that represent the same spatial region receive identical rotations, thereby preserving relative positional structure both within and across streams. However, positional information alone cannot distinguish tokens from different streams, as the $|\mathcal{P}|$ tokens associated with the same spatial location are otherwise indistinguishable to the attention operator. We therefore augment each compact token with a learned scale-identity embedding, specific to each block and stream, before attention. Ablating this embedding degrades FID from 2.31 to 3.29 (Appendix~\ref{app:scale_identity}), demonstrating the importance of explicitly encoding stream identity.

Each stream applies its own query, key, value, and output projections to its compact tokens. The resulting queries, keys, and values are concatenated along the sequence dimension and processed jointly by a single attention operator:
\begin{equation}
\begin{aligned}
    &\bm{Q}=[\bm{Q}_p]_{p\in\mathcal{P}},\quad
    \bm{K}=[\bm{K}_p]_{p\in\mathcal{P}},\quad
    \bm{V}=[\bm{V}_p]_{p\in\mathcal{P}},\\
    &\bm{O}=\operatorname{softmax}\!\left(
    \frac{\bm{Q}\bm{K}^\top}{\sqrt{d_h}}\right)\bm{V},
\end{aligned}
    \label{eq:joint_attention}
\end{equation}
using $D/d_h$ heads over a sequence of length $|\mathcal{P}|M$. The output is then split by stream, passed through the corresponding output projection, expanded via $\bm{W}^{\mathrm{e}}_p$, and added to each stream through a zero-initialized gate. Crucially, no cross-scale attention mask is imposed. In contrast to prior pixel-space architectures that restrict information flow from coarse to fine~\citep{yu2026pixeldit,ma2026deco,dip2026,wang2026pixnerd,he2026hyperdit}, joint attention permits unrestricted bidirectional interaction among all active scales. A single softmax over the concatenated sequence therefore allows the model to adaptively determine the extent and direction of cross-scale information exchange at each layer. Section~\ref{sec:flow_analysis} shows that enforcing scale-causal attention severely degrades performance.

\begin{table*}[t]
\centering
\caption{Model configurations.  Depths are listed from the finest to the
coarsest stream.  Because the streams are right-aligned, the total number of
blocks equals the largest per-stream depth.}
\vspace{-1mm}
\label{tab:model_configs}
\setlength{\tabcolsep}{9pt}
\begin{tabular}{@{}lcccccc@{}}
\toprule
\textbf{Model} & \textbf{Patch sizes} & \textbf{Depths} & \textbf{Stream widths} & $D$ & \textbf{Heads} & \textbf{Params} \\ \midrule
\ours-L   & $(1,4,16)$ & $(4,8,22)$  & $(16,256,1152)$ & 1152 & 16 & 788M \\
\ours-XL  & $(1,4,16)$ & $(4,12,26)$ & $(16,256,1152)$ & 1152 & 16 & 963M \\
\ours-H   & $(1,2,4,8,16)$ & $(4,8,12,20,26)$ & $(16,64,256,512,1152)$ & 1152 & 16 & 1.4B \\
\ours-T2I & $(1,4,16)$ & $(2,6,14)$  & $(16,256,1536)$ & 1536 & 24 & 1.4B \\ \bottomrule
\end{tabular}
\end{table*}

\subsection{Architectural Details}
\label{sec:t2i_method}

Viewing scales as distinct modalities naturally extends the architecture to text conditioning. We project precomputed text features to width $D$, assign them 1D rotary positional embeddings, and concatenate them with the compact image tokens when constructing $\bm{Q}$, $\bm{K}$, and $\bm{V}$ in Equation~\ref{eq:joint_attention}. The text stream is active in every block and, analogous to each image stream, maintains its own attention projections, adaptive normalization, and feed-forward network. Image generation remains entirely in pixel space: the output is decoded from the $1\times1$ stream, without an image autoencoder.

Following prior work~\citep{yu2025representation}, we add a REPA loss to
Equation~\ref{eq:flow_matching}, aligning the coarsest-stream tokens with frozen
DINOv2 features~\citep{oquab2024dinov2} of the clean image, with weight
$\lambda=0.5$, at block 8 for class-conditional models and block 6 for
text-to-image models. The remaining architectural components follow standard practice. Queries and keys are normalized with RMSNorm~\citep{zhang2019root,dehghani2023scaling}, and feed-forward networks use SwiGLU~\citep{shazeer2020glu}. Each stream maintains its own modulation parameters, applied at the native stream resolution before compaction and following the attention residual.

\begin{table*}[t]
\centering
\caption{Comparison of pixel-space text-to-image models at $512\times512$.  We
report the per-category GenEval scores, the overall GenEval score, and the average DPG-Bench score.}
\label{tab:t2i}
\setlength{\tabcolsep}{6pt}
\begin{tabular}{lccccccccc}
\toprule
& & \multicolumn{7}{c}{\textbf{GenEval$\uparrow$}}
& \textbf{DPG$\uparrow$} \\
\cmidrule(lr){3-9} \cmidrule(lr){10-10}
\textbf{Method}
& \textbf{Parameters}
& \shortstack{\textbf{Single}\\\textbf{Object}}
& \shortstack{\textbf{Two}\\\textbf{Objects}}
& \textbf{Counting}
& \textbf{Colors}
& \textbf{Position}
& \shortstack{\textbf{Color}\\\textbf{Attribution}}
& \textbf{Overall}
& \textbf{Average} \\
\midrule
PixelFlow
& 0.9B & -- & -- & -- & -- & -- & -- & 0.60 & 77.9 \\
PixNerd
& 1.2B & 0.97 & 0.86 & 0.44 & 0.83 & 0.71 & 0.53 & 0.73 & 80.9 \\
PixelGen
& 1.1B & 0.99 & 0.88 & 0.59 & 0.90 & 0.70 & 0.70 & 0.79 & -- \\
PixelDiT
& 1.3B & \textbf{1.00} & \textbf{0.94} & 0.70 & 0.90 & 0.53 & 0.65 & 0.78 & 83.7 \\
DeCo
& 1.1B & \textbf{1.00} & 0.92 & \textbf{0.72} & 0.91 & \textbf{0.80} & \textbf{0.79} & \textbf{0.86} & 81.4 \\
\rowcolor{oursgreen}\textbf{\ours}
& 1.4B & 0.99 & \textbf{0.94} & \textbf{0.72} & \textbf{0.94} & 0.78 & 0.77 & \textbf{0.86} & \textbf{83.9} \\
\bottomrule
\end{tabular}
\end{table*}

\section{Experiments}
\vspace{-1mm}

We evaluate \ours{} on class-conditional ImageNet
generation~\citep{deng2009imagenet} and on text-to-image generation.  On
ImageNet, we report both system-level comparisons with prior work and
controlled architectural ablations.  Following standard practice, we report
FID, sFID, Inception Score
(IS), precision, and
recall; lower values are better for FID and
sFID, and higher values are better for the remaining metrics.  For
text-to-image generation, we assess compositional and prompt-following ability
at $512\times512$ resolution using GenEval~\citep{ghosh2023geneval} and
DPG-Bench~\citep{hu2024ella}.

\begin{table}[t]
\centering
\caption{Class-conditional generation on ImageNet $512\times512$.  All metrics
are computed from 50K samples generated with classifier-free guidance.  Bold denotes the best pixel-space result.}
\label{tab:imagenet_512}
\setlength{\tabcolsep}{1.6pt}
\renewcommand{\arraystretch}{1.04}
\begin{tabular}{lccccc}
\toprule
\textbf{Method} & \textbf{FID}$\downarrow$ & \textbf{sFID}$\downarrow$ & \textbf{IS}$\uparrow$ & \textbf{Prec.}$\uparrow$ & \textbf{Rec.}$\uparrow$ \\ \midrule
\rowcolor{sectiongray}\multicolumn{6}{l}{\textit{Latent-space diffusion}} \\
        
DiT-XL~\citep{peebles2023scalable} & 3.04 & 5.02 & 240.8 & 0.84 & 0.54 \\
SiT-XL~\citep{ma2024sit} & 2.62 & 4.18 & 252.2 & 0.84 & 0.57 \\
MaskDiT~\citep{zheng2024fast} & 2.50 & 5.10 & 256.2 & 0.83 & 0.56 \\
U-ViT-H~\citep{bao2023all} & 4.05 & -- & 263.8 & 0.84 & 0.48 \\
REPA~\citep{yu2025representation} & 2.08 & 4.19 & 274.6 & 0.83 & 0.58 \\
RAE-XL~\citep{zheng2025diffusion} & 1.13 & -- & 259.6 & 0.80 & 0.63 \\ \midrule
\rowcolor{sectiongray}
\multicolumn{6}{l}{\textit{Pixel-space diffusion}} \\
ADM~\citep{dhariwal2021diffusion} & 3.85 & 5.86 & 221.7 & \textbf{0.84} & 0.53 \\
RIN~\citep{jabri2023scalable} & 3.95 & -- & 216.0 & -- & -- \\
Simple Diffusion~\citep{hoogeboom2023simple} & 3.54 & -- & 205.0 & -- & -- \\
VDM++~\citep{kingma2023understanding} & 2.65 & -- & 278.1 & -- & -- \\
PixNerd-XL~\citep{wang2026pixnerd} & 2.84 & 5.95 & 245.6 & 0.80 & 0.59 \\
EPG-L/32~\citep{liu2026there} & 2.35 & -- & 295.4 & 0.82 & 0.57 \\
DeCo-XL/16~\citep{ma2026deco} & 2.22 & \textbf{4.67} & 290.0 & 0.80 & 0.60 \\
JiT-H~\citep{li2026back} & 1.94 & -- & 309.1 & -- & -- \\
PixelU-H/32~\citep{guo2026pixelu} & 1.92 & -- & \textbf{322.1} & -- & -- \\
PixelDiT~\citep{yu2026pixeldit} & 1.81 & 5.61 & 278.6 & 0.78 & \textbf{0.67} \\
\rowcolor{oursgreen}\textbf{\ours} & \textbf{1.44} & 4.76 & 319.2  &  0.80 & 0.64 \\ \bottomrule
\end{tabular}

\end{table}

\vspace{-1mm}
\subsection{Experimental Setup}
\label{sec:settings}

We train directly on RGB pixels, without a VAE or learned image tokenizer.
Class- and text-conditional models share the optimizer, initialization, and
classifier-free guidance dropout, and differ only in the condition stream,
data, and resolution schedule.  On ImageNet, ablations use the $(1,4,16)$
streams and are evaluated after 80 epochs; \ours{}-L and \ours{}-XL are
trained for 600 epochs and \ours{}-H for 360.  Metrics are computed on 50K
samples drawn with a flow DPM solver.  The text-to-image model follows the
three-stage recipe and data of DeCo~\citep{ma2026deco}.  Architectures, learning rates,
guidance intervals, and the full sampling protocol are given in
Appendix~\ref{app:hparams}.

Table~\ref{tab:model_configs} lists the configurations used for the main
results.  All models share a joint-attention patch size of $a=16$, a per-head
dimension of 72 (class-conditional) or 64 (text-conditional), SwiGLU
feed-forward networks with an expansion ratio of 4, RMSNorm on queries and
keys, and 2D rotary position embeddings on the compact grid.

\subsection{Class-Conditional Generation}

\paragraph{Results at $256\times256$.}
Table~\ref{tab:merged_imagenet_256} compares our models with latent- and
pixel-space generators.  Among pixel-space methods, \ours{}-L improves upon
the reproduced PixelDiT-XL baseline from 1.66 to 1.55 FID at a comparable
inference parameter budget.  It also attains higher recall (0.65 vs.\ 0.64)
and substantially lower sFID (4.42 vs.\ 4.89), suggesting more faithful
spatial structure rather than merely sharper textures.  For \ours{}-XL, increasing the depth further reduces FID to 1.50 and sFID to 4.31, with an IS of 308.7.
\ours{}-H extends the model to all five scales $(1,2,4,8,16)$ and achieves 1.43
FID and 4.13 sFID after only 360 epochs.  This result mirrors the scale-wise scaling law, with additional scales continuing to
improve the model.  Relative to the strongest pixel-space baselines, \ours{}-H improves FID by $0.18$ over
PixelDiT-XL and by $0.20$ over PixelU-H/16.

\vspace{-3mm}
\paragraph{Results at $512\times512$.}
Table~\ref{tab:imagenet_512} summarizes the comparison at higher resolution.
The advantage of multi-scale streams increases with resolution.  We
hypothesize that higher resolutions require modeling structure over a wider
range of spatial scales, which a sequence of intermediate granularities
captures more effectively than a two-level hierarchy.  \ours{} achieves 1.44 FID and 319.2 IS, outperforming PixelDiT (1.81),
PixelU-H/32 (1.92), and JiT-H (1.94) with a large margin. Figures~\ref{fig:qual_c2i_256}
and~\ref{fig:qual_c2i_512} present samples at both resolutions, and
additional samples from other classes are provided in Appendix~\ref{app:qualitative}.

\vspace{-1mm}
\subsection{Text-to-Image Generation}

The same multi-scale interface accommodates a text stream without any change
to the pixel readout.  As shown in Table~\ref{tab:t2i}, \ours{} attains a
GenEval score of 0.86 at $512\times512$, matching the best pixel-space result,
reported by DeCo-XXL/16, and outperforming PixelDiT-T2I (0.78) and
PixNerd-XXL/16 (0.73).  On DPG-Bench, it achieves 83.9, exceeding DeCo by 2.5 points and PixelDiT-T2I by 0.2
points despite a shorter training schedule.  These results indicate that treating text as an additional modality alongside
several image granularities is a viable approach to pixel-space text-to-image
generation.  Figure~\ref{fig:qual_t2i} presents
$512\times512$ samples, and the corresponding prompts are listed in
Appendix~\ref{app:qualitative}.

\begin{figure*}[t]
\centering

\begin{subfigure}{\textwidth}
    \centering
    \includegraphics[width=\linewidth]{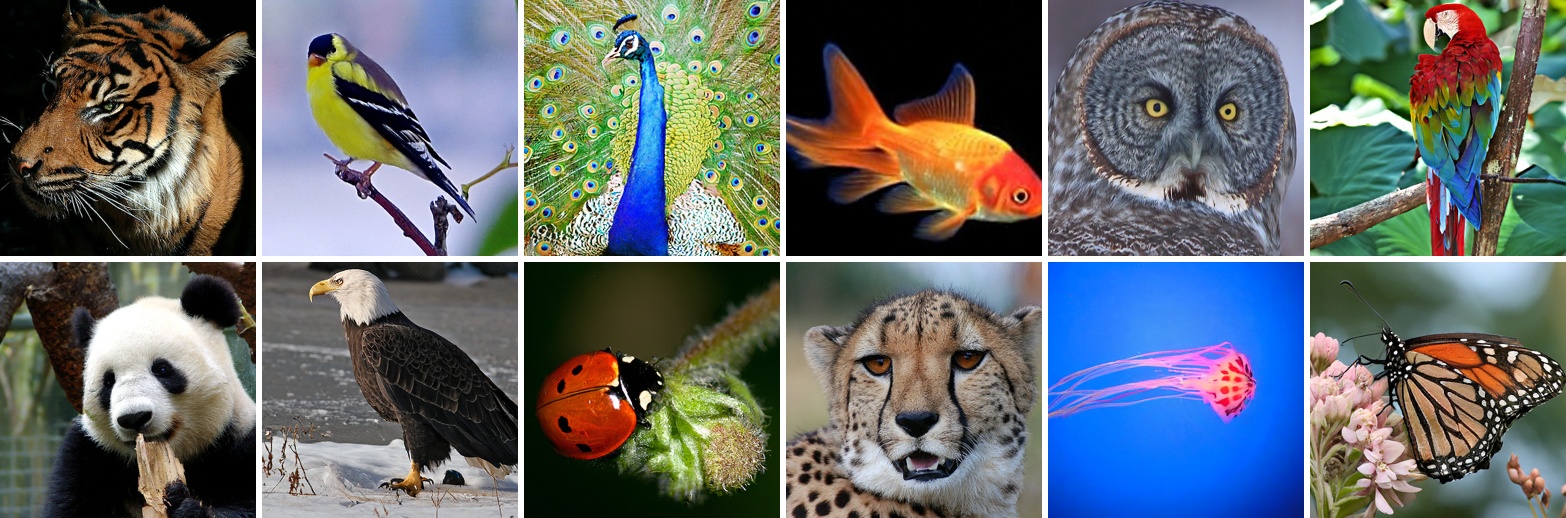}
    \caption{Class-conditional samples from \ours{}-XL on ImageNet
    $256\times256$ (CFG scale 4.0).}
    \label{fig:qual_c2i_256}
\end{subfigure}

\vspace{0.5em}

\begin{subfigure}{\textwidth}
    \centering
    \includegraphics[width=\linewidth]{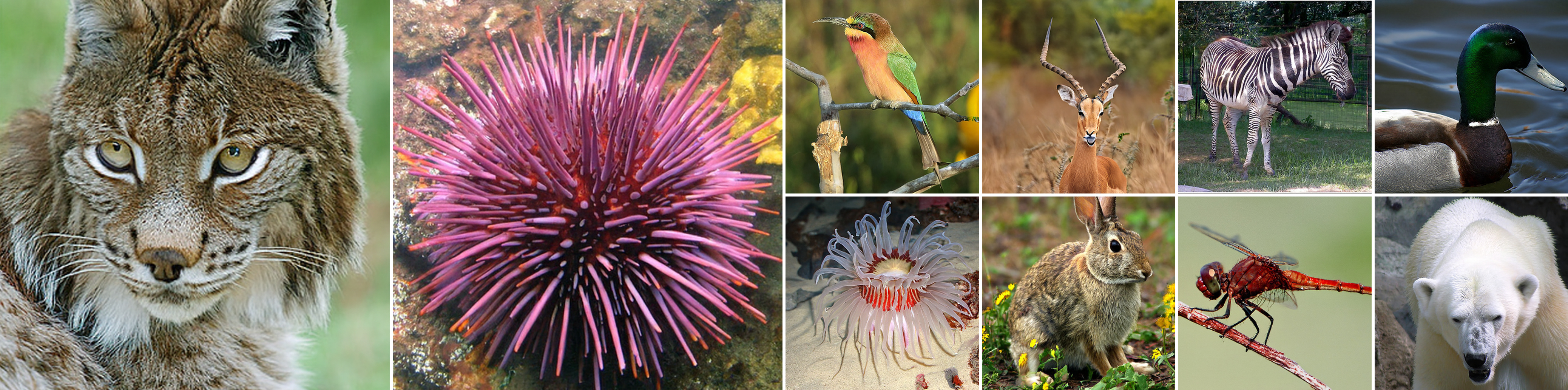}
    \caption{Class-conditional samples from \ours{} on ImageNet
    $512\times512$ (CFG scale 4.0).}
    \label{fig:qual_c2i_512}
\end{subfigure}

\vspace{0.5em}

\begin{subfigure}{\textwidth}
    \centering
    \includegraphics[width=\linewidth]{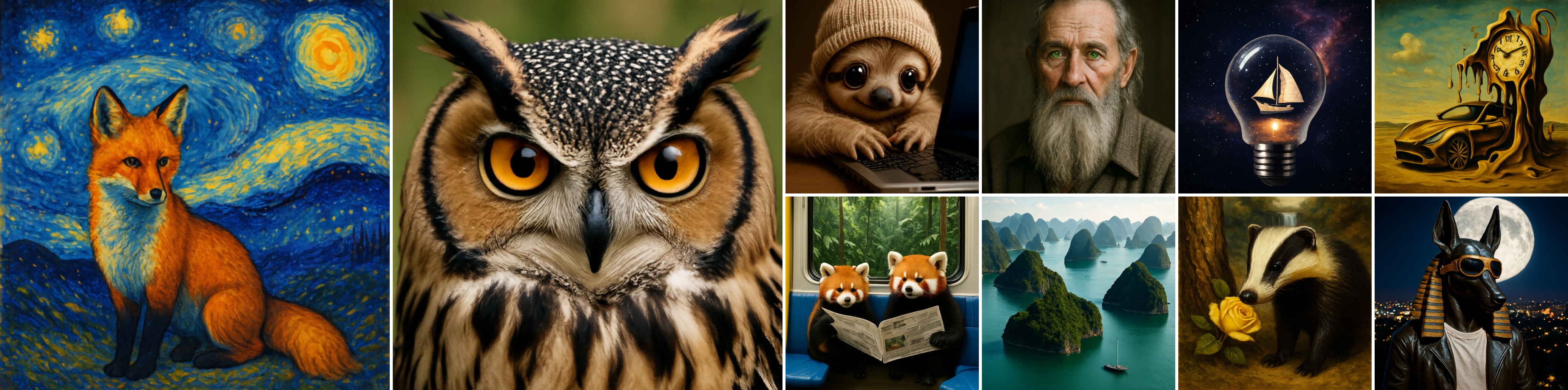}
    \caption{Text-to-image samples from \ours{} at $512\times512$.
    Prompts are drawn from DrawBench and PartiPrompts and are listed,
    from left to right, in Table~\ref{tab:t2i_prompts}.}
    \label{fig:qual_t2i}
\end{subfigure}

\caption{Qualitative generation results of \ours{} on
class-conditional and text-to-image generation tasks.}
\label{fig:qualitative_results}

\end{figure*}

\section{Analysis}

Unless otherwise stated, all analysis experiments use the 788M-parameter
$(1,4,16)$ model trained for 80 epochs on ImageNet $256\times256$, with the same
sampling and evaluation protocol as the main comparison.

\subsection{Scale-Wise Scalability}

\label{sec:scale_scaling}

Table~\ref{tab:scale_scaling} examines whether introducing additional scale-specific computation benefits a model that already contains a coarse $16\times16$ stream and a pixel-level $1\times1$ stream. Adding an eight-block $4\times4$ stream reduces FID from 4.39 to 2.31 and increases IS from 241.5 to 302.1. By comparison, allocating a similar parameter budget to additional coarse-stream depth yields a substantially smaller improvement (Section~\ref{sec:scale_choice}). These results support our central empirical finding: the number of represented granularities provides a distinct and effective axis for scaling model capacity. Across the evaluated configurations, FID generally improves as more scales are introduced: from 4.39 with two scales to 2.31 with three, 2.26 with four $(1,2,4,16)$, and 2.19 with all five $(1,2,4,8,16)$. Performance therefore does not saturate within the range of scales considered, suggesting further headroom along the scale dimension at the current training budget.

\begin{table*}[t]
\centering
\caption{Scale-wise model scaling on ImageNet $256\times256$.  Each scale
column gives the number of transformer blocks assigned to the corresponding
patch stream, and a dash indicates that the stream is absent.  FID improves as
scales are added, and the five-scale model performs best.}
\label{tab:scale_scaling}
\setlength{\tabcolsep}{9pt}
\begin{tabular}{@{}ccccc|cc|ccccc@{}}
\toprule
\multicolumn{5}{c|}{Stream Depth} & \multicolumn{2}{l|}{Training Settings} & \multicolumn{5}{c}{Quality} \\ \midrule
16$\times$16 & 8$\times$8 & 4$\times$4 & 2$\times$2 & 1$\times$1 & \multicolumn{1}{l}{Params} & Epochs & FID$\downarrow$ & IS$\uparrow$ & sFID$\downarrow$ & Prec.$\uparrow$ & Rec.$\uparrow$ \\ \midrule
22 & - & - & - & 4 & 651M & 80 & 4.39  & 241.5 & 6.23 & 0.73 & \textbf{0.59} \\
22 & - & 8 & - & 4 & 788M & 80 & 2.31 & \textbf{302.1} & 4.77 & \textbf{0.81} & 0.58 \\
22 & - & 8 & 8 & 4 & 910M & 80 & 2.26 & 298.5 & \textbf{4.66} & \textbf{0.81} & 0.58 \\
22 & 8 & 8 & - & 4 & 914M & 80 & 2.35 & 300.0 & 4.67 & \textbf{0.81} & \textbf{0.59} \\
22 & 8 & 8 & 8 & 4 & 1.0B & 80 & \textbf{2.19} & 296.4 & 4.73 & \textbf{0.81} & \textbf{0.59} \\ \bottomrule
\end{tabular}
\end{table*}

\subsection{Choice of Scales}
\label{sec:scale_choice}

The preceding comparison varies both the number of scales and the parameter count. Table~\ref{tab:scale_choice} therefore controls for model size: all five configurations contain 758--788M parameters and share the same $1\times1$ output stream. Allocating the additional capacity solely to the coarse stream, 26 blocks at $16\times16$ together with the pixel stream achieves an FID of 3.63, while distributing depth uniformly across three scales (13 blocks per stream) performs slightly worse, with an FID of 3.74. In contrast, every multi-scale configuration with an asymmetric depth allocation substantially outperforms both baselines. These results indicate that the gain observed in Table~\ref{tab:scale_scaling} arises primarily from introducing an intermediate granularity rather than merely increasing parameter count. Among the candidate intermediate scales, $4\times4$ performs best, achieving 2.31 FID and 4.77 sFID, compared with 2.42 FID for $8\times8$ and 2.54 for $2\times2$.

\begin{table*}[t]
\centering
\caption{Controlled comparison of intermediate patch scales on ImageNet
$256\times256$.  Models have comparable parameter counts and share a
$1\times1$ output stream and a $16\times16$ backbone stream.  The
$4\times4$ intermediate stream gives the best FID, IS, and sFID.}
\label{tab:scale_choice}
\setlength{\tabcolsep}{8pt}
\begin{tabular}{@{}ccccc|cc|ccccc@{}}
\toprule
\multicolumn{5}{c|}{Stream Depth} & \multicolumn{2}{l|}{Training Settings} & \multicolumn{5}{c}{Quality} \\ \midrule
16$\times$16 & 8$\times$8 & 4$\times$4 & 2$\times$2 & 1$\times$1 & \multicolumn{1}{l}{Params} & Epochs & FID$\downarrow$ & IS$\uparrow$ & sFID$\downarrow$ & Prec.$\uparrow$ & Rec.$\uparrow$ \\ \midrule
\rowcolor{oursgreen}22 & - & 8 & - & 4 & 788M & 80 & \textbf{2.31} & \textbf{302.1} & \textbf{4.77} &\textbf{ 0.81} & 0.58 \\
13 & - & 13 & - & 13 & 768M & 80 & 3.74 & 237.8  &5.63 & 0.76 & \textbf{0.59} \\
26 & - & - & - & 4 & 758M & 80 & 3.63 & 254.6 & 5.72 & 0.76 &  0.58 \\
22 & - & - & 8 & 4 & 773M & 80 & 2.54 & 292.5 & 5.26 & 0.80 & 0.58 \\
22 & 8 & - & - & 4 & 777M & 80 & 2.42 & 298.2 & 5.12 & 0.80  & \textbf{0.59} \\ \bottomrule
\end{tabular}

\end{table*}

\subsection{Cross-Scale Information Flow}
\label{sec:flow_analysis}

Our default joint attention permits bidirectional information flow across scales, marking a key departure from prior pixel-space diffusion architectures. In PixelDiT, DeCo, DiP, PixNerd, and HyperDiT, information flows predominantly from a coarse semantic pathway to a finer pathway, without a corresponding fine-to-coarse interaction. If such a strictly top-down inductive bias were sufficient, explicitly enforcing it should preserve or improve performance.

Table~\ref{tab:mask_anchor} tests this hypothesis using two cross-scale attention masks. The scale-causal mask allows a query at scale $p$ to attend only to keys at scales $p'\ge p$, permitting fine streams to access coarser streams while preventing the reverse interaction. The strict variant additionally removes same-scale attention from all but the coarsest stream, yielding a purely top-down information pathway. With the default late anchoring, the scale-causal mask degrades FID from 2.31 to 3.47, while the strict mask yields 2.55. Under middle-layer anchoring, the corresponding FIDs are 2.50 and 3.24. In both settings, unrestricted bidirectional attention achieves the best performance.

These results suggest that fine-to-coarse communication provides useful information beyond what is available through a purely top-down pathway. Because coarse tokenization discards local spatial detail, fine streams can expose information unavailable to the $16\times16$ representation alone, allowing coarse-scale features to incorporate finer-grained evidence during joint attention. Such interactions cannot be represented by a strictly coarse-to-fine architecture. We therefore retain fully bidirectional joint attention across active scales.

\begin{table}[]
\centering
\caption{Effect of cross-scale information flow and fine-stream anchor location on ImageNet $256\times256$. ``Last Layers'' denotes the default right-aligned schedule, and ``Middle Layers'' places the fine-stream blocks in the middle of the coarse backbone. The scale-causal mask allows a query to attend only to its own or coarser scales, and the strict variant additionally removes same-scale attention except in the coarsest stream.}
\label{tab:mask_anchor}
\setlength{\tabcolsep}{1.4pt}
\begin{tabular}{@{}lccccc@{}}
\toprule
\textbf{Method} & \textbf{FID$\downarrow$} & \textbf{IS$\uparrow$} & \textbf{sFID$\downarrow$} & \textbf{P$\uparrow$} & \textbf{R$\uparrow$} \\ \midrule
Last Layers (Ours) & \textbf{2.31} & \textbf{302.1} & \textbf{4.77} & \textbf{0.81} & 0.58 \\
+ Scale Causal Mask & 3.47 & 262.2 & 5.58 & 0.76 & 0.58 \\
+ Strict Scale Causal Mask & 2.55 & 299.4 & 5.18 & 0.79 & \textbf{0.59} \\ \midrule
Middle Layers & \textbf{2.31} & 277.4 & 4.99 & 0.80 & \textbf{0.59} \\
+ Scale Causal Mask & 2.50 & 282.6 & 5.10 & 0.80 & 0.58 \\
+ Strict Scale Causal Mask & 3.24 & 265.2 & 5.70 & 0.78 & 0.58 \\ \bottomrule
\end{tabular}
\end{table}

\subsection{Where to Anchor the Fine Streams}

\label{sec:anchor_analysis}

We next examine where the limited fine-stream depth should be allocated relative to the coarse backbone. Right-aligning the fine streams in the final layers achieves 2.31 FID and 302.1 IS. Moving the same number of fine-stream blocks to intermediate depths, with individual fine-stream layers interleaved with several coarse-backbone layers as in HyperDiT, yields a similar FID but lowers IS to 277.4 and worsens sFID to 4.99. These results favor late anchoring as the default design. Under this configuration, the coarse stream first develops a global representation, after which the fine streams are introduced to jointly refine local appearance and produce the final pixel-level output.

\section{Conclusion}

We introduced \ours{}, a multi-scale pixel-space diffusion transformer that exploits flexible pixel tokenization by treating spatial scale as a modality. By representing the same noisy image at multiple granularities and coupling scale-specific streams through compacted joint attention, \ours{} enables efficient and bidirectional information exchange across scales while generating images directly in pixel space. Our experiments establish the number of scales as an effective architectural scaling dimension, with additional granularities providing substantial gains beyond simply increasing network depth. Further analysis highlights the importance of bidirectional cross-scale communication and late fine-stream computation, suggesting that effective pixel-space generation benefits from first establishing global structure and subsequently integrating fine-grained information through joint refinement. Across ImageNet class-conditional and text-to-image generation, \ours{} demonstrates strong performance and scales naturally to higher resolutions and additional conditioning modalities. These results suggest that multi-scale representation provides a promising architectural direction for scalable pixel-space generative modeling.

\subsection*{LLM Usage Statement}

In this work, we used generative AI tools to assist with experiment-related coding, as well as drafting, editing, and polishing the manuscript. Generative AI tools were not used to generate the core research ideas, hypotheses, or methodological contributions of this work. All AI-assisted code was reviewed, tested, and validated by the authors. The authors take full responsibility for the final content of this work, including all text, claims, code, experimental results, and other artifacts produced with the aid of generative AI.

{
    \small
    \bibliographystyle{ieeenat_fullname}
    \bibliography{main}
}

\appendix

\begin{table*}[h]
\centering
\caption{Per-stage text-to-image training settings.  The global batch size is
the product of the per-GPU batch size, the number of GPUs, and the gradient
accumulation factor (always one).  Steps denote the number of additional
optimizer updates in each stage.}
\label{tab:t2i_stages}
\setlength{\tabcolsep}{6pt}
\renewcommand{\arraystretch}{1.08}
\begin{tabular}{lccccccc}
\toprule
\textbf{Stage} & \textbf{Res.} & \textbf{Data} & \textbf{Steps} &
\textbf{GPUs} & \textbf{Global batch} & \textbf{LR} & \textbf{Shift} \\ \midrule
1. Pretrain    & $256^2$ & BLIP3o        & 200K & 24 & 1536 & $2\times10^{-4}$ & 2.0 \\
2. High-res.   & $512^2$ & BLIP3o ($\geq$512px) & 80K & 32 & 512 & $1\times10^{-4}$ & 4.0 \\
3. SFT         & $512^2$ & BLIP3o-60k    & 40K  & 32 & 512  & $1\times10^{-5}$ & 4.0 \\
\bottomrule
\end{tabular}
\end{table*}

\begin{table*}[t]
\centering
\caption{Effect of the joint-attention patch size $a$ on ImageNet
$256\times256$ after 80 epochs.  Larger attention patches shorten the joint
sequence but coarsen cross-scale correspondence.}
\label{tab:attn_patch_size}
\begin{tabular}{@{}l|c|ccccc@{}}
\toprule
\textbf{Attention Patch Size} & \textbf{Epochs} & \textbf{FID$\downarrow$} & \textbf{IS$\uparrow$} & \textbf{sFID$\downarrow$} & \textbf{Prec.$\uparrow$} & \textbf{Rec.$\uparrow$} \\ \midrule
16 (ours) & 80 & \textbf{2.31} & \textbf{302.1} & \textbf{4.77} & \textbf{0.81} & 0.58 \\
32 & 80 & 3.00  & 252.6  & 5.53 & 0.77 & \textbf{0.60} \\ \bottomrule
\end{tabular}
\end{table*}

\section{Extended Related Work}
\label{app:related}

\subsection{Latent Diffusion and Diffusion Transformers}
Latent diffusion moves denoising into the compressed space of a pretrained
autoencoder~\citep{rombach2022high}, and DiT established the transformer as the
standard backbone in this space~\citep{peebles2023scalable}; it was
subsequently extended to flow matching by SiT~\citep{ma2024sit} and to text
conditioning by MM-DiT~\citep{esser2024scaling}.  Much subsequent work targets
the autoencoder itself.  VA-VAE aligns latents with vision foundation model
features~\citep{yao2025reconstruction}, REPA-E tunes the autoencoder and the
diffusion model jointly~\citep{leng2025repae}, and representation autoencoders
replace the variational bottleneck with frozen self-supervised
features~\citep{zheng2025diffusion}.  REPA shows that aligning intermediate
denoiser features with DINOv2 substantially accelerates
training~\citep{yu2025representation,oquab2024dinov2}, and DDT decouples
semantic encoding from detail decoding within the latent
space~\citep{wang2025ddt}.

\subsection{Pixel-Space Diffusion}
Diffusion models were originally formulated in pixel
space~\citep{sohl2015deep,ho2020denoising,song2021score,dhariwal2021diffusion},
where high resolutions were first reached through cascades of separately
trained super-resolution
stages~\citep{ho2022cascaded,saharia2022photorealistic,teng2024relay} or
through architectures that concentrate computation on a small set of latent
tokens~\citep{jabri2023scalable}.  U-Net and U-ViT variants with aggressive
early downsampling and long skip connections remain competitive at high
resolution~\citep{hoogeboom2023simple,hoogeboom2024simpler,kingma2023understanding}.
The recent generation of single-stage pixel transformers is more directly
related to our work.  PixelFlow uses a single set of parameters but traverses
cascading resolutions during sampling~\citep{chen2025pixelflow}.  JiT argues
that predicting clean data rather than a noised quantity is what makes large
pixel patches tractable~\citep{li2026back}, and EPG attains strong FID by
pretraining the encoder with a self-supervised objective prior to generative
fine-tuning~\citep{liu2026there}.  PixelGen adds LPIPS and DINO perceptual
supervision on top of
$x$-prediction~\citep{ma2026pixelgen}.

Most closely related to our setting are models that explicitly partition the
network by granularity.  PixelDiT runs a $16\times16$ patch-level DiT and a
$1\times1$ pixel-level DiT, connects them through pixel-wise adaptive
normalization, and compacts pixel tokens before attention~\citep{yu2026pixeldit}.
DeCo replaces the fine pathway with an attention-free pixel decoder and
reweights the flow-matching loss with JPEG-derived frequency
priors~\citep{ma2026deco}.  DiP employs a convolutional patch detailer
head~\citep{dip2026}, and PixNerd predicts per-patch neural-field weights from
the coarse stream~\citep{wang2026pixnerd}.  HyperDiT lets a small-patch stream
query multi-level semantic anchors of a large-patch stream through
cross-attention, using scale-aware rotary embeddings to align the two
coordinate systems~\citep{he2026hyperdit}.  In each case, there are exactly two
granularities, conditioning runs strictly from coarse to fine, and the coarse
stream never attends to the fine one.  PixelU instead argues that such decoders
are unnecessary under $x$-prediction and replaces them with skip connections
around a downsampled bottleneck~\citep{guo2026pixelu}; this removes the decoder
but retains a single tokenization and a two-resolution hierarchy.  In contrast,
we employ several granularities, assign each a complete transformer stream,
and couple them bidirectionally within a single attention operation.

\subsection{Multi-Scale Representations}
Multi-scale representations have long been central to computer vision, from feature pyramids to hierarchical vision transformers~\citep{lin2017feature,liu2021swin}. In generative modeling, scale is also widely exploited, but is typically organized \emph{sequentially}: cascaded diffusion models generate low-resolution images before refining them at higher resolutions~\citep{ho2022cascaded}, while visual autoregressive models progressively generate images from coarse to fine scales~\citep{tian2024visual}. Other approaches, such as Matryoshka diffusion, jointly model nested resolutions within a shared architecture~\citep{gu2024matryoshka}. In contrast, \ours{} uses multiple scales as concurrent representations within a single forward pass. All streams observe the same image at the same resolution and noise level, with scale controlling \emph{how the signal is tokenized} rather than \emph{when it is generated}.

\section{Detailed Settings and Hyperparameters}
\label{app:hparams}

\subsection{Initialization and Shared Optimization}

Patch-embedding and compaction matrices are initialized with Xavier
initialization, whereas the adaptive-normalization and output projections are
zero-initialized.  Consequently, every residual block and every cross-scale
gate is initialized as the identity mapping, and the streams become coupled
only as training progresses.  Both class- and text-conditional models optimize
with AdamW at a constant learning rate, clip
gradients to unit norm, maintain an exponential moving average (EMA) of the
weights for evaluation, and drop the condition with probability $0.1$ to
enable classifier-free guidance (CFG).  They differ only in the condition
stream, data pipeline, resolution schedule, and the per-task hyperparameters
below.

\subsection{Class-Conditional Training and Sampling}

Class-conditional models are trained directly on RGB images from the
ImageNet-1K training set, without a VAE or any learned image tokenizer.
Unless otherwise specified, ablations use the $(1,4,16)$ streams with widths
$(16,256,1152)$ and depths $(4,8,22)$, matching \ours{}-L in
Table~\ref{tab:model_configs}.  Training is performed in bfloat16 with a
global batch size of 256 on eight GPUs, using AdamW
($\beta=(0.9,0.999)$, learning rate $10^{-4}$, no weight decay), gradient
clipping at 1.0, and an EMA decay of 0.9999.  \ours{}-L and \ours{}-XL are
trained for 600 epochs and \ours{}-H for 360 epochs, whereas ablation models
are evaluated after 80 epochs.  Timesteps are sampled as
$t=\mathrm{sigmoid}(\epsilon)$ with $\epsilon\sim\mathcal{N}(0,1)$ and then
time-shifted, with a shift of 1.0 at $256\times256$ and 3.0 at $512\times512$.
Class labels are dropped with probability 0.1 for classifier-free guidance.
REPA uses DINOv2 ViT-B/14 features, a three-layer MLP projector, alignment at
block 8, and $\lambda=0.5$.

For evaluation at $256\times256$, we generate 50K samples with a 100-step
flow DPM solver.  The 80-epoch ablations use a CFG scale of 3.25 applied over
$t\in[0.1,1.0]$, whereas the main models use a CFG scale of 2.75 applied over
$t\in[0.1,0.9]$.  At $512\times512$, we initialize from the
$256\times256$ checkpoint, fine-tune with a reduced learning rate and a
shifted time schedule, and sample with a CFG scale of 3.5.  All
distributional metrics are computed against the standard ADM ImageNet
reference statistics~\citep{dhariwal2021diffusion}.

\begin{table*}[h]
\centering
\caption{Ablation of the learned scale-identity embedding on ImageNet
$256\times256$.  Scale-identity embeddings distinguish streams that share the
same compact attention coordinates; removing them degrades FID from 2.31 to
3.29.}
\label{tab:scale_identity}
\begin{tabular}{@{}l|c|ccccc@{}}
\toprule
\textbf{Method} & \textbf{Epochs} & \textbf{FID$\downarrow$} & \textbf{IS$\uparrow$} & \textbf{sFID$\downarrow$} & \textbf{Prec.$\uparrow$} & \textbf{Rec.$\uparrow$} \\ \midrule
\ours & 80 & \textbf{2.31} & \textbf{302.1} & \textbf{4.77} & \textbf{0.81} & 0.58 \\
w/o Scale Identity Embedding & 80 & 3.29 & 265.2  & 6.02 & 0.76 &\textbf{0.59} \\ \bottomrule
\end{tabular}
\end{table*}

\begin{table*}[t]
\centering
\caption{Effect of the frequency-aware flow-matching (FreqFM) loss on ImageNet
$256\times256$.  Adding FreqFM does not improve any of the evaluated metrics
after 80 epochs.}
\label{tab:freqfm}
\begin{tabular}{@{}l|c|ccccc@{}}
\toprule
\textbf{Method} & \textbf{Epochs} & \textbf{FID$\downarrow$} & \textbf{IS$\uparrow$} & \textbf{sFID$\downarrow$} & \textbf{Prec.$\uparrow$} & \textbf{Rec.$\uparrow$} \\ \midrule
\ours & 80 & \textbf{2.31} & \textbf{302.1} & \textbf{4.77} & \textbf{0.81} & \textbf{0.58} \\
+ $\mathcal{L}_\text{FreqFM}$ & 80 & 2.40  & 300.0  & 4.95 & \textbf{0.81} &  \textbf{0.58}\\ \bottomrule
\end{tabular}
\end{table*}
\begin{table*}[t]
\centering
\caption{Ablation of representation alignment (REPA) on ImageNet
$256\times256$.  ``Largest Patch'' aligns the compact spatial tokens of the
$16\times16$ stream with frozen DINOv2 features, whereas ``Registers'' aligns
learned, non-spatial register tokens instead.  All models are evaluated after
80 epochs.}
\label{tab:repa_ablation}
\begin{tabular}{@{}l|c|ccccc@{}}
\toprule
\textbf{Method} & \textbf{Epochs} & \textbf{FID$\downarrow$} & \textbf{IS$\uparrow$} & \textbf{sFID$\downarrow$} & \textbf{Prec.$\uparrow$} & \textbf{Rec.$\uparrow$} \\ \midrule
Largest Patch (Ours) & 80 & \textbf{2.31} & 302.1 & 4.77 & 0.81 & \textbf{0.58} \\
No REPA & 80 & 5.30 & 195.3 & 5.63 & \textbf{0.83} &  0.47 \\
Registers & 80 & 2.56 & \textbf{312.5} & \textbf{4.43} & \textbf{0.83} & 0.56 \\ \bottomrule
\end{tabular}
\end{table*}

\subsection{Text-to-Image Training}
\label{app:t2i}

The text-conditional model uses image streams $(1,4,16)$ with per-stream widths
$(16,256,1536)$ and depths $(2,6,14)$, a joint width of $D=1536$ with 24 heads,
and an attention patch size of 16, identical to that of the class-conditional
models.  Pixels are read out from the $1\times1$ stream.  Text is encoded by a
frozen Qwen3-1.7B model~\citep{yang2025qwen3}, projected from 2048 channels to
the joint width, and assigned its own RoPE frequencies with $\theta=10^4$.  We
use a fixed 128-token context without an attention mask, so padding tokens
remain visible to the image streams.  Text embeddings are dropped with
probability 0.1 for classifier-free guidance.

Training proceeds in the three stages listed in Table~\ref{tab:t2i_stages},
which follow the public recipe of DeCo~\citep{ma2026deco}.  Per-GPU batch
sizes, learning rates, and data are taken unchanged from its released
configurations, so that the comparison in Table~\ref{tab:t2i} isolates the
effect of the architecture.  The high-resolution stage is shorter than that
of DeCo (80K vs.\ 100K steps); all other settings are identical.  Stages~1
and~2 train on the BLIP3o pretraining
corpus~\citep{chen2025blip3o}, with stage~2 restricted to images whose shorter
side is at least 512 pixels, and stage~3 fine-tunes on BLIP3o-60k.  Each stage
resumes from the previous checkpoint and continues the cumulative step count.
Steps are counted as optimizer updates without gradient accumulation.

All stages use AdamW without weight decay, a constant learning rate, gradient
clipping at 1.0, bfloat16 mixed precision with attention accumulated in fp32,
gradient checkpointing, and an EMA decay of 0.9999.  The Adam betas are
$(0.9,0.95)$ for the two pretraining stages and $(0.9,0.999)$ for supervised
fine-tuning.  Timesteps are sampled from a logit-normal distribution with mean
0 and standard deviation 1 and then time-shifted by the factor given in
Table~\ref{tab:t2i_stages}.  The loss combines the flow-matching objective
with representation alignment at block 6 (weight 0.5, DINOv2 ViT-B/14 targets).

\paragraph{Evaluation protocol.}
All reported text-to-image results are obtained with the final stage-3 EMA
weights and a second-order flow DPM solver with 50 sampling steps, a CFG scale
of 4.0 applied over the full time interval, and a time shift of 3.0, generating
at $512\times512$ in bfloat16.  For GenEval~\citep{ghosh2023geneval}, we sample
four images for each of the 553 official prompts with seed 0 and score them
with the official detector pipeline; the reported value is the mean over the
six task categories.  For DPG-Bench~\citep{hu2024ella}, we sample four images
for each of the 1065 prompts with seed 0 and score them with the official ELLA
implementation.

\section{Additional Analysis}

\subsection{Scale Identity Embedding}
\label{app:scale_identity}

Because all streams are compacted onto the same attention grid and share the
same rotary coordinates, position alone does not indicate the tokenization
from which a token originates.  Table~\ref{tab:scale_identity} shows that the
learned scale-identity embedding is necessary: removing it degrades FID from
2.31 to 3.29 and sFID from 4.77 to 6.02.  This embedding is the multi-scale
counterpart of the modality embeddings used in multimodal transformers.

\subsection{Attention Patch Size}

The attention patch size $a$ determines the length of the joint sequence and,
consequently, the granularity at which streams exchange information.  Doubling
$a$ to 32 reduces the number of attention tokens by a factor of four, but it
also quadruples the number of native tokens that each compact token must
summarize.  Table~\ref{tab:attn_patch_size} shows that this trade-off is
unfavorable: FID degrades from 2.31 to 3.00 and IS from 302.1 to 252.6.  We
therefore use $a=16$ throughout, a choice that also renders the compaction of
the coarse stream an identity mapping in the spatial dimension.

\subsection{Frequency-Aware Flow-Matching Loss}
\label{app:freqfm}

DeCo reweights the flow-matching residual in the DCT domain using JPEG
quantization tables, on the grounds that a single network would otherwise
expend capacity on perceptually irrelevant high frequencies~\citep{ma2026deco}.
We evaluated the same loss on top of our model.  As shown in
Table~\ref{tab:freqfm}, it yields no improvement: FID increases from 2.31 to
2.40 and sFID from 4.77 to 4.95, while precision and recall remain unchanged.
We interpret this result as mild support for our framing.  Frequency
reweighting informs a single-scale network which components of the residual
are perceptually important; when separate streams already specialize by
granularity, such guidance is largely redundant.

\subsection{Representation Alignment}
\label{sec:repa_analysis}

Table~\ref{tab:repa_ablation} shows that representation alignment remains
important for optimization in pixel space, even with multiple streams.
Removing REPA increases FID from 2.31 to 5.30 and reduces IS from 302.1 to
195.3 after 80 epochs. Replacing these spatial tokens with learned
register tokens improves IS and sFID but worsens FID to 2.56.  We therefore
align the spatial tokens of the coarse stream in the main model.

\subsection{Computational Cost}

Table~\ref{tab:flops_256} reports the parameter count and per-forward compute
of pixel-space diffusion models at $256\times256$.  \ours{}-L requires 186
GFLOPs, which is 30 GFLOPs (19\%) more than PixelDiT-XL at a nearly identical
parameter count (788M vs.\ 797M), while improving FID from 1.66 to 1.55 and
sFID from 4.89 to 4.42 (Table~\ref{tab:merged_imagenet_256}).  Relative to
JiT-G/16, \ours{}-L uses less than half the compute and 40\% of the parameters
yet attains a lower FID (1.55 vs.\ 1.82).  It is also roughly $16\times$
cheaper than PixelFlow-XL/4 and $6\times$ cheaper than ADM-G, which process
long token sequences or dense convolutional feature maps at high resolution.

\begin{table}[t]
\centering
\caption{Parameters and compute of pixel-space diffusion models on ImageNet
$256\times256$.  GFLOPs are measured for a single forward pass without
classifier-free guidance, counting one multiply--add as one FLOP.}
\label{tab:flops_256}
\setlength{\tabcolsep}{12pt}
\renewcommand{\arraystretch}{1.08}
\begin{tabular}{lcc}
\toprule
\textbf{Method} & \textbf{Params} & \textbf{GFLOPs} \\ \midrule
ADM-G~\citep{dhariwal2021diffusion}          & 554M & 1120\\
RIN~\citep{jabri2023scalable}                & 410M & 334 \\
VDM++~\citep{kingma2023understanding}        & 2B   & 555\\
PixelFlow-XL/4~\citep{chen2025pixelflow}     & 677M & 2909\\
PixNerd-XL/16~\citep{wang2026pixnerd}        & 700M & 134 \\
JiT-G/16~\citep{li2026back}                  & 2B   & 383 \\
PixelDiT-XL~\citep{yu2026pixeldit}           & 797M & 156 \\
\rowcolor{oursgreen}\textbf{\ours-L}         & 788M & 186 \\
\bottomrule
\end{tabular}
\end{table}

\section{Qualitative Examples}
\label{app:qualitative}

\begin{figure*}[h]
\centering
\includegraphics[width=\textwidth]{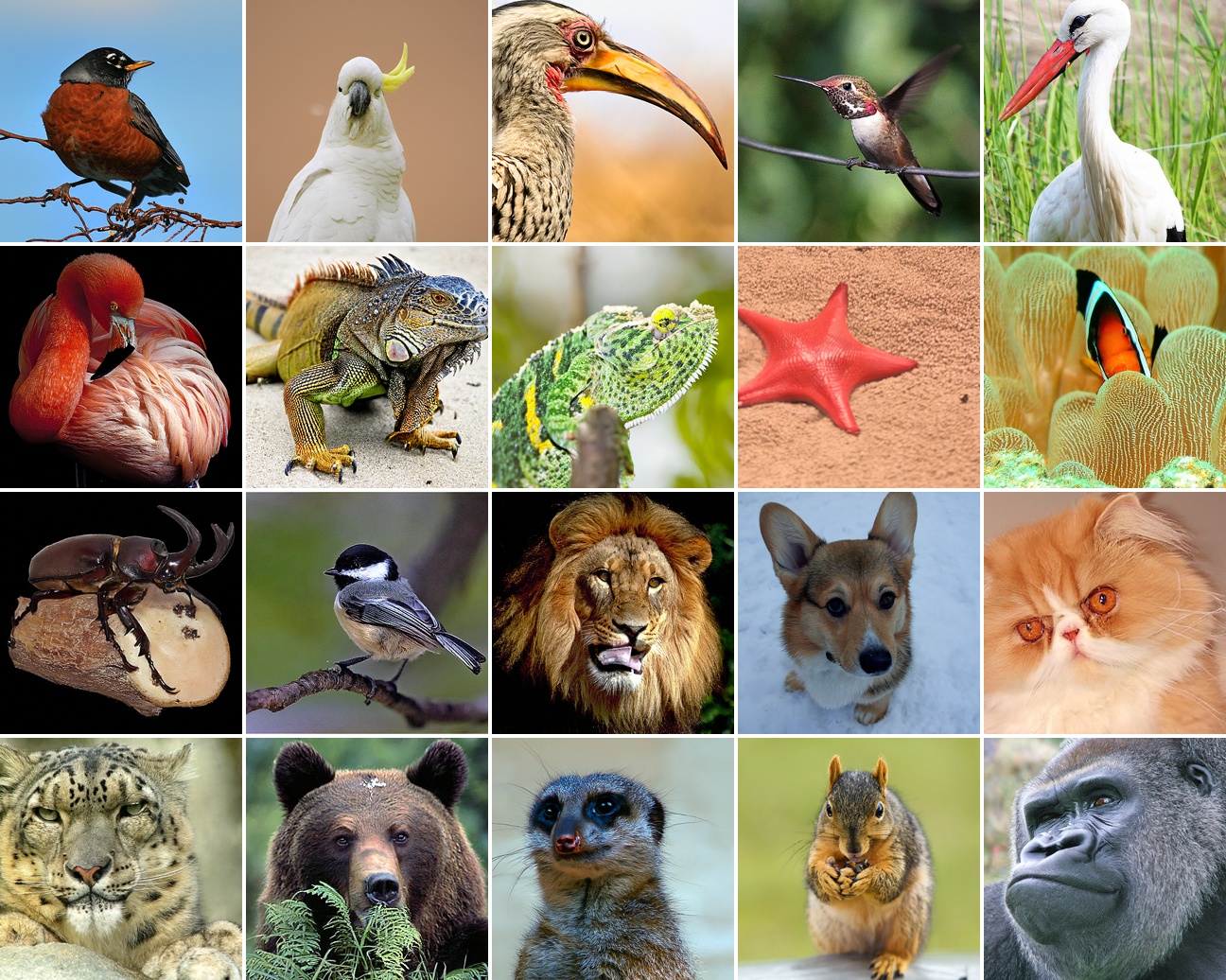}
\caption{Class-conditional samples from \ours{}-XL on ImageNet $256\times256$.
Classes from left to right, top to bottom: robin (15), sulphur-crested cockatoo (89),
hornbill (93), hummingbird (94), white stork (127), flamingo (130), common iguana (39),
African chameleon (47), starfish (327), anemone fish (393), rhinoceros beetle (306),
chickadee (19), lion (291), Pembroke corgi (263), Persian cat (283), snow leopard (289),
brown bear (294), meerkat (299), fox squirrel (335), and gorilla (366).}
\label{fig:app_c2i_256}
\end{figure*}

\begin{figure*}[h]
\centering
\includegraphics[width=\textwidth]{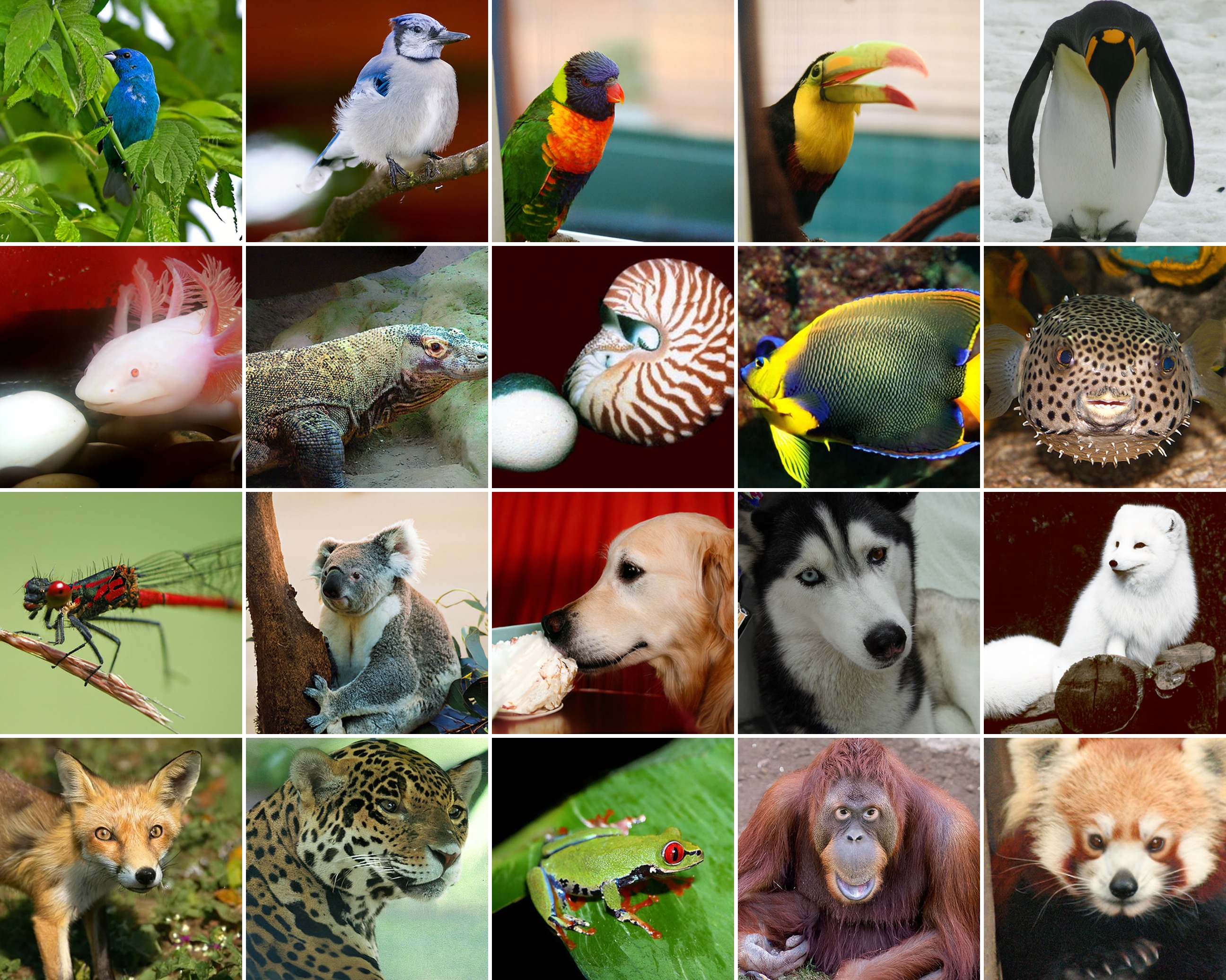}
\caption{Class-conditional samples from \ours{} on ImageNet $512\times512$.
Classes from left to right, top to bottom: indigo bunting (14), jay (17), lorikeet (90),
toucan (96), king penguin (145), axolotl (29), Komodo dragon (48), chambered nautilus (117),
rock beauty (392), puffer (397), damselfly (320), koala (105), golden retriever (207),
Siberian husky (250), Arctic fox (279), red fox (277), jaguar (290), tree frog (31),
orangutan (365), and lesser panda (387).}
\label{fig:app_c2i_512}
\end{figure*}

\begin{figure*}[h]
\centering
\includegraphics[width=\textwidth]{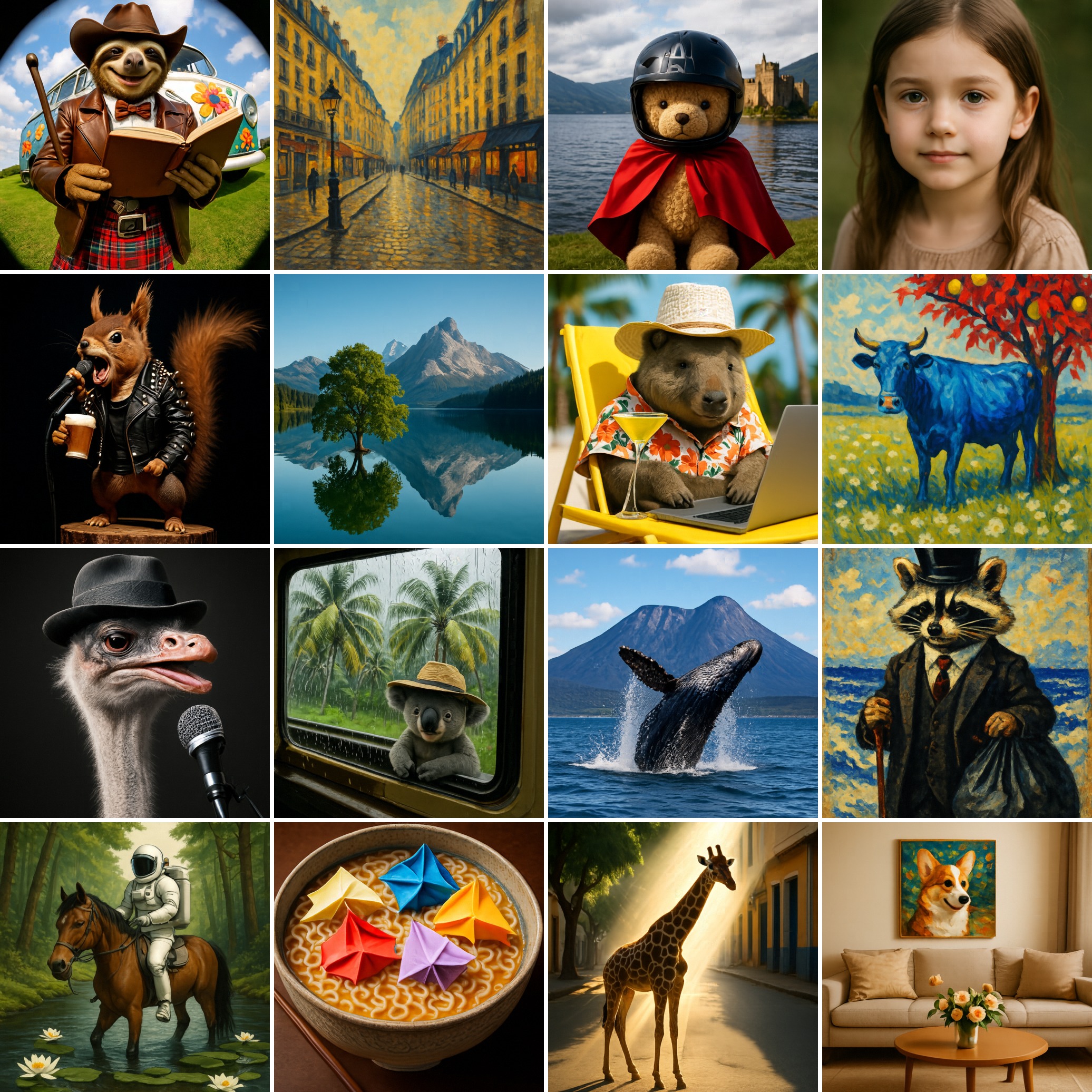}
\caption{Additional text-to-image samples from \ours{} at $512\times512$.
Prompts are taken from PartiPrompts and are listed, from left to right and top
to bottom, in Table~\ref{tab:app_t2i_prompts}.}
\label{fig:app_t2i_512}
\end{figure*}

Figures~\ref{fig:app_c2i_256} and~\ref{fig:app_c2i_512} present one sample for
each of twenty ImageNet classes at $256\times256$ (\ours{}-XL) and at
$512\times512$ (\ours{}), all generated with a CFG scale of 4.0. The classes are
disjoint from those shown in Figures~\ref{fig:qual_c2i_256} and~\ref{fig:qual_c2i_512}.
Table~\ref{tab:t2i_prompts} lists the prompts for the text-to-image samples in
Figure~\ref{fig:qual_t2i}. Figure~\ref{fig:app_t2i_512} shows sixteen
additional $512\times512$ text-to-image samples, covering photorealistic
scenes, portraits, landscapes, painterly styles, and multi-object compositional
prompts from PartiPrompts; their prompts are listed in
Table~\ref{tab:app_t2i_prompts}.

\begin{table*}[h]
\centering
\caption{Prompts for Figure~\ref{fig:qual_t2i}, ordered from left to right.}
\label{tab:t2i_prompts}
\small
\begin{tabular}{@{}r p{0.9\textwidth}@{}}
\toprule
\# & Prompt \\
\midrule
1 & a painting of a fox in the style of starry night. \\
2 & a close-up of the eyes of an owl. \\
3 & A fluffy baby sloth with a knitted hat trying to figure out a laptop,
close up, highly detailed, studio lighting, screen reflecting in its eyes. \\
4 & an old man with a long grey beard and green eyes. \\
5 & A photo of a light bulb in outer space traveling the galaxy with a
sailing boat inside the light bulb. \\
6 & a sport car melting into a clock, surrealist painting in the style of
Salvador Dali. \\
7 & A photograph of the inside of a subway train. There are red pandas
sitting on the seats. One of them is reading a newspaper. The window shows the
jungle in the background. \\
8 & An aerial view of Ha Long Bay. \\
9 & A richly textured oil painting of a young badger delicately sniffing a
yellow rose next to a tree trunk. A small waterfall can be seen in the
background. \\
10 & a portrait of a statue of the Egyptian god Anubis wearing aviator
goggles, white t-shirt and leather jacket. A full moon over the city of Los
Angeles is in the background at night. \\
\bottomrule
\end{tabular}
\end{table*}

\begin{table*}[h]
\centering
\caption{Prompts for Figure~\ref{fig:app_t2i_512}, ordered from left to right
and top to bottom.}
\label{tab:app_t2i_prompts}
\small
\begin{tabular}{@{}r p{0.9\textwidth}@{}}
\toprule
\# & Prompt \\
\midrule
1 & A smiling sloth is wearing a leather jacket, a cowboy hat, a kilt and a
bowtie. The sloth is holding a quarterstaff and a big book. The sloth is
standing on grass a few feet in front of a shiny VW van with flowers painted on
it. wide-angle lens from below. \\
2 & a painting of street in Paris \\
3 & A teddy bear wearing a motorcycle helmet and cape is standing in front of
Loch Awe with Kilchurn Castle behind him \\
4 & a portrait of young girl \\
5 & A punk rock squirrel in a studded leather jacket shouting into a
microphone while standing on a stump and holding a beer on dark stage. \\
6 & Snow mountain and tree reflection in the lake \\
7 & A wombat sits in a yellow beach chair, while sipping a martini that is on
his laptop keyboard. The wombat is wearing a white panama hat and a floral
Hawaiian shirt. Out-of-focus palm trees in the background. \\
8 & a blue cow is standing next to a tree with red leaves and yellow fruit.
the cow is standing in a field with white flowers. impressionistic painting \\
9 & a photograph of an ostrich wearing a fedora and singing soulfully into a
microphone \\
10 & A train ride in the monsoon rain in Kerala. With a Koala bear wearing a
hat looking out of the window. There is a lot of coconut trees out of the
window. \\
11 & a whale breaching near a mountain \\
12 & A raccoon wearing formal clothes, wearing a tophat and holding a cane.
The raccoon is holding a garbage bag. Oil painting in the style of Hokusai. \\
13 & A photo of an astronaut riding a horse in the forest. There is a river in
front of them with water lilies. \\
14 & A high resolution photo of a large bowl of ramen. There are several
origami boats in the ramen of different colors. \\
15 & a light shining on a giraffe in a street \\
16 & A cozy living room with a painting of a corgi on the wall above a couch
and a round coffee table in front of a couch and a vase of flowers on a coffee
table \\
\bottomrule
\end{tabular}
\end{table*}

\end{document}